\documentclass{article}

\usepackage{microtype}
\usepackage{graphicx}
\usepackage{subfigure}
\usepackage{booktabs} 
\usepackage{amsmath}
\usepackage{balance}
\usepackage{hyperref}

\usepackage[accepted]{icml2021}

\icmltitlerunning{Submission and Formatting Instructions for ICML 2021}

\begin{document}

\twocolumn[
\icmltitle{Interpretable Face Manipulation Detection via Feature Whitening}



\icmlsetsymbol{corresponding}{*}

\begin{icmlauthorlist}
\icmlauthor{Yingying Hua}{1,2}
\icmlauthor{Daichi Zhang}{1,2}
\icmlauthor{Pengju Wang}{1}
\icmlauthor{Shiming Ge}{1,2,corresponding}
\end{icmlauthorlist}

\icmlaffiliation{1}{Institute of Information Engineering, Chinese Academy of Sciences, Beijing 100195, China}
\icmlaffiliation{2}{School of Cyber Security, University of Chinese Academy of Sciences, Beijing 100049, China}

\icmlcorrespondingauthor{Shiming Ge}{geshiming@iie.ac.cn}

\icmlkeywords{Interpretability, Face Manipulation Detection, Feature Whitening.}

\vskip 0.3in
]



\printAffiliationsAndNotice{}  

\begin{abstract}
Why should we trust the detections of deep neural networks for manipulated faces?
Understanding the reasons is important for users in improving the fairness, reliability, privacy and trust of the detection models.
In this work, we propose an interpretable face manipulation detection approach to achieve the trustworthy and accurate inference.
The approach could make the face manipulation detection process transparent by embedding the feature whitening module.
This module aims to whiten the internal working mechanism of deep networks through feature decorrelation and feature constraint.
The experimental results demonstrate that our proposed approach can strike a balance between the detection accuracy and the model interpretability.
\end{abstract}

\section{Introduction}
Face manipulation detection~\cite{rossler2019faceforensics,guarnera2020deepfake} aims to discriminate between the real and fake facial images. There have been many detection methods proposed with the increasing interest in face manipulation~\cite{tolosana2020deepfakes}. Although the detection models perform well, why should we trust them and how they come to the predictions? It is quite important to understand the underlying reasons~\cite{molnar2020interpretable,gilpin2018explaining,arrieta2020explainable} behind the model predictions for face manipulation detection. We expect the model to answer the "Why?" and "How?" questions, ensuring that users could evaluate the detection results.

The interpretability research~\cite{zhang2020a,zhang2018visual} towards face manipulation detection has proposed some technical methods to understand what properties of facial images make them detectable.
These interpretable methods~\cite{chai2020what,trinh2021interpretable} mostly focus on the relation between the manipulated faces and the detection results, ignoring the internal working mechanism of deep models. In this way, they could only answer the "Why?" question, but stay confused about how it comes to a particular detection. In addition, the interpretability research is still insufficient and in need of further improvements on face manipulation detection. To address this challenging issue, this paper aims to explain face manipulation detection by making the inference process transparent from the perspective of human beings.

Inspired by whitening that uses a decorrelation transformation~\cite{kessy2018optimal,chen2020concept}, we propose a feature whitening (FW) module to reduce complexity and improve understandability inside the black-box deep models.
The FW module aims to achieve the decorrelation and constraint of feature representations without affecting the detection accuracy.
We propose a decorrelation algorithm based on the zero-phase components analysis (ZCA) whitening~\cite{huang2019iterative,huangi2018decorrelated} to obtain the decorrelated feature activations, and then establish the constraints between these decorrelated activations and the manipulated faces.
In other words, the FW module just reorganizes the internal feature representations for better understanding, and it hardly damages the working mechanism of deep networks. Thus, the detection performance will hardly be influenced while improving the interpretability of face manipulation detection.

Therefore, we develop an interpretable face manipulation detection (IFMD) approach by embedding the FW module into a deep network.
The FW module can provide explanations on the underlying working mechanism of deep models, which decorrelates the latent features to distribute them into different channels and constrains the decorrelated activations to represent the manipulated faces.
Furthermore, it is a structurally independent module with a wide range of applications.
Thus, we can embed the FW module into different layers of a deep network to reduce the complexity and improve the readability for the internal representations.
In this way, our approach is able to discriminate a manipulated facial image accurately, and provide explanations on how it comes to the detection and why should we trust it. The IFMD approach can facilitate the development and deployment of face manipulation detection models in critical environments in the long term.

The main contributions of our paper are as follows. 1) We propose an interpretable face manipulation detection approach to explain the internal mechanism of face manipulation detection models, that can strike a balance between the model interpretability and detection accuracy.
2) We design a feature whitening module to whiten the face manipulation detection process through the decorrelation and constraint of the underlying feature representations inside deep models.
3) We conduct experiments to verify the effectiveness of our approach, demonstrating that the IFMD approach could provide the trustworthy and accurate detection.

\section{Approach}

\subsection{Problem Formulation}
For the given samples $\boldsymbol{X}=\{\boldsymbol{x}_1, \boldsymbol{x}_2,..., \boldsymbol{x}_n\}$ and their corresponding labels $Y=\{y_1, y_2,..., y_n\}$, we aim to make the face manipulated detection process transparent for better understanding. In order to achieve this, we propose the feature whitening algorithm to deconstruct the internal representations, including the feature decorrelation and feature constraint. Considering the model complexity, the feature representations could be distributed across different channels respectively through feature decorrelation. Then, we constrain the decorrelated representations mapping the input to improve the model readability.

Given the feature representations $\boldsymbol{z}_i=f(\boldsymbol{x}_i)$ extracted with a feature extractor $f$, the output of the FW module $\phi(\cdot)$ is shown as follows:
\begin{equation}\label{1}
  \hat{\boldsymbol{z}_i}=\phi(\boldsymbol{z}_i),
\end{equation}
where $\hat{\boldsymbol{z}_i}$ is the whitened representations.
Then, we feed the whitened representations into a classifier $g$ for prediction.
The cross-entropy loss function of the IFMD model is defined as
\begin{equation}\label{2}
  \mathcal L = \sum_{i=1}^n g(\phi(f(\boldsymbol{x}_i; \omega); \mu), y_i; \nu),
\end{equation}
where $\omega$ is the parameter of the feature extractor for the input, $\mu$ is the parameter of the FW module, $\nu$ is the parameter of the classifier for the whitened features.

The FW module aims to whiten the underlying mechanism of face manipulation detection models and consists of two operations, including feature decorrelation and feature constraint.
The details are as follows.

\subsection{Feature Decorrelation}
The purpose of the feature decorrelation is to reduce the complexity of the latent features inside deep models.
We use the ZCA algorithm~\cite{kessy2018optimal,huang2019iterative} based on iterative normalization to achieve feature decorrelation.
The decorrelation algorithm can be viewed as a generalization of standardizing a random variable that converts a random vector with mean and covariance matrix into a new one.
For a batch representations $\boldsymbol{Z}$ of size $m$, the decorrelated representations $\boldsymbol{Z}_d$ of are calculated by
\begin{equation}\label{3}
  \boldsymbol{Z}_d = d(\boldsymbol{Z})=\boldsymbol{D}(\boldsymbol{Z}-\frac{1}{m}\boldsymbol{Z}\cdot\boldsymbol{1}\cdot\boldsymbol{1}^T),
\end{equation}
where $d(\cdot)$ is the decorrelation operation, $\boldsymbol{D}$ is the decorrelation matrix.
Then we calculate the decorrelation matrix $\boldsymbol{D}=\boldsymbol{\Sigma}^{-1/2}=[\boldsymbol{d}] \rm diag(\sigma)^{-1/2}[\boldsymbol{d}]^T$.
And $[\boldsymbol{d}]$ and $\rm diag(\sigma)$ are the eigenvalues and associated eigenvectors of $\boldsymbol{\Sigma}$.
The covariance matrix is defined as
\begin{equation}\label{4}
  \boldsymbol{\Sigma}=\frac{1}{m}(\boldsymbol{Z}-\frac{1}{m}\boldsymbol{Z}\cdot\boldsymbol{1}\cdot\boldsymbol{1}^T)(\boldsymbol{Z}-\frac{1}{m}\boldsymbol{Z}\cdot\boldsymbol{1}\cdot\boldsymbol{1}^T)^T.
\end{equation}
To improve the optimization efficiency and detection performance, we calculate the covariance matrix based on one feasible transformation $\boldsymbol{\Sigma}_N=\boldsymbol{\Sigma}/\sqrt{tr(\boldsymbol{\Sigma})}$ as follows:
\begin{equation}\label{5}
  \boldsymbol{\Sigma}^{-1/2}=\boldsymbol{P}/\sqrt{tr(\boldsymbol{\Sigma})},
\end{equation}
where $\boldsymbol{P}$ is the inverse square root, and $tr(\cdot)$ denotes the matrix trace.
To avoid executing eigen-decomposition or SVD, we approximately calculate the square root inverse by Newton's iteration~\cite{bini2005algorithms}.
Based on the Newton's method, we can update the inverse square root at each step as follows:
\begin{equation}\label{6}
\begin{aligned}
  \begin{cases}
    \boldsymbol{P}_0 = \boldsymbol{I}\\
    \boldsymbol{P}_t = \frac{1}{2}(3\boldsymbol{P}_{t-1}-\boldsymbol{P}_{t-1}^3\boldsymbol{\Sigma}_N)\\
\end{cases}
\end{aligned},
\end{equation}
where $t$ is the training epoch.
In this way, the decorrelation algorithm can reduce the internal correlation of feature representations and distribute them into different channels for better understanding.

\subsection{Feature Constraint}
The feature constraint aims to constrain the decorrelated representations mapping the original input space to improve the semantic interpretability.
Given the decorrelated representations $\boldsymbol{Z}_d$, we apply the constraint operation $c(\cdot)$ to further whiten them as follows:
\begin{equation}\label{7}
  \hat{\boldsymbol{Z}}=c(\boldsymbol{Z}_d)=\boldsymbol{C}^T\boldsymbol{Z}_d,
\end{equation}
where $\boldsymbol{C}$ is the constraint matrix.
In order to calculate the constraint matrix, we use the generalized Cayley transformation~\cite{huang2018recursive} method for efficient computation. During the training process, the constraint matrix can be updated by
\begin{equation}\label{8}
\begin{aligned}
  \begin{cases}
    \boldsymbol{C}_0 = \boldsymbol{I}\\
    \boldsymbol{C}_t = (1+\frac{\lambda}{2})\boldsymbol{S}^{-1}(1-\frac{\lambda}{2})\boldsymbol{S}\boldsymbol{C}_{t-1}\\
\end{cases}
\end{aligned},
\end{equation}
where $\lambda$ is the learning rate updated by curvilinear search at each optimization step,
$\boldsymbol{S}$ is the skew-symmetric matrix based on the stochastic gradient $\boldsymbol{G}$ of the loss function.
The skew-symmetric matrix can be defined as
\begin{equation}\label{9}
  \boldsymbol{S} = \boldsymbol{G}\boldsymbol{C}_{t-1}^T - \boldsymbol{C}_{t-1}\boldsymbol{G}^T.
\end{equation}
We calculate and update the gradient of the loss function with a mini-batch of samples at each step.
We use the exponential moving average to update this parameter more smoothly as follows:
\begin{equation}\label{10}
\begin{aligned}
  \begin{cases}
    \boldsymbol{G}_0 = \boldsymbol{\nabla}{\boldsymbol{C}}\\
    \boldsymbol{G}_t = \alpha\boldsymbol{G}_{t-1}+(1-\alpha)\boldsymbol{G}_t\\
\end{cases}
\end{aligned},
\end{equation}
where $\boldsymbol{\nabla}$ is the gradient of the loss function $\mathcal L$ with respect to the constraint matrix, and $\alpha$ is the decay factor.
The feature constraint guides the decorrelated features to represent the input's semantics for greater readability.

Based on the above analysis, the FW module applies the feature decorrelation and constraint operations to whiten the black-box deep models.
We embed the FW module into an original detection model at different layers to obtain the IFMD model.
In addition, we apply some training tricks to improve the optimization process.

\section{Experiments}
In order to verify our approach, we train different IFMD models based on the original models, including Xception, ResNet18, VGG16-BN.
And then we compare their performance of detection and interpretability.
The dataset consists of the real faces from FFHQ and the fake faces generated by StyleGAN~\cite{karras2020a}.
We train these deep models with the parameters of epoch 40, momentum 0.9, weight decay $5\times10^{-4}$, learning rate 0.1.

\begin{table}[h]
\caption{The accuracy ($\%$) of face manipulation detection. FW-L means that the FW module is embedded into the low layers of the original models. Similarly, FW-M and FW-H means the middle and high layers. The IFMD models also perform well with respect to the detection accuracy.}
\renewcommand\arraystretch{1.3}
\label{table1}
\vskip 0.12in
\begin{center}
\begin{tabular}{cccc}
\toprule
Model & Xception & ResNet18 & VGG16-BN \\
\midrule
None      & 99.76 & \textbf{99.41} & 99.28 \\
FW-L    & \textbf{99.89} & 99.09 & \textbf{99.63} \\
FW-M & 98.60 & 99.30 & 99.13 \\
FW-H   & 99.90 & 99.17 & 98.81 \\
\bottomrule
\end{tabular}
\end{center}
\vskip -0.1in
\end{table}

\begin{figure}[t]
\vskip 0.15in
\begin{center}
\centerline{\includegraphics[width=\columnwidth]{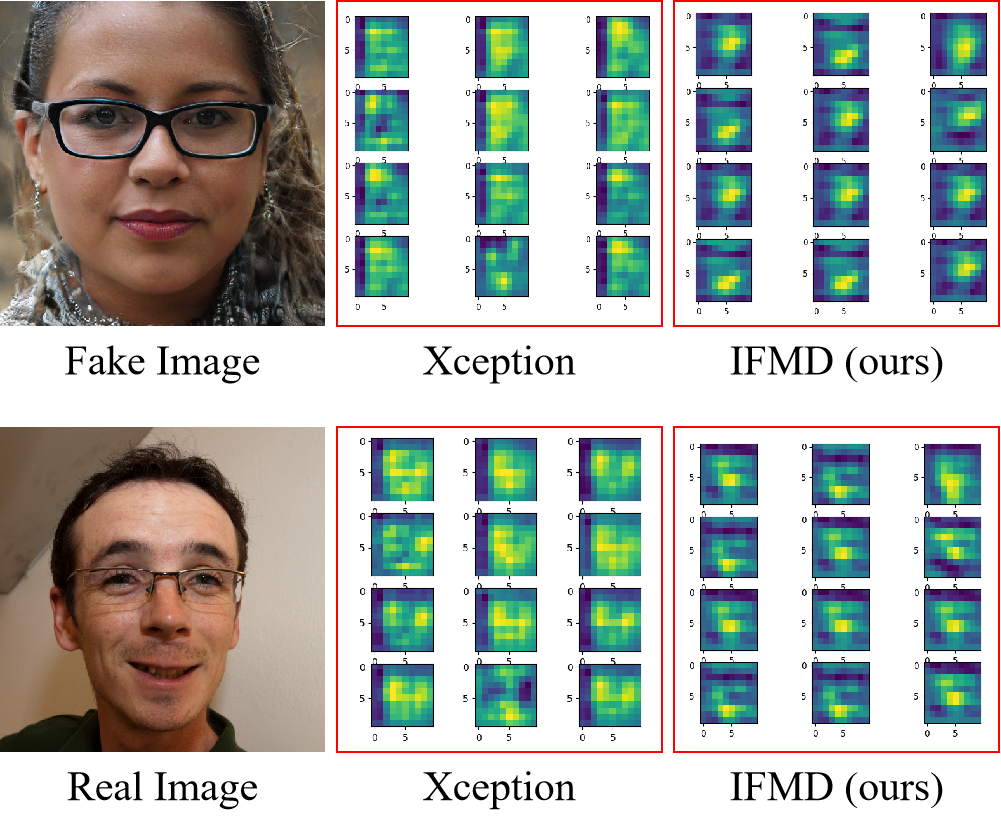}}
\caption{The feature maps of the second last layer across different channels. Compared with the Xception model, we find that the feature maps of our IFMD model are more salient and their high-level semantic information is easier to understand for people.}
\label{fig1}
\end{center}
\vskip -0.25in
\end{figure}

\subsection{Detection Performance}

Considering the detection performance, we compare the accuracy of the IFMD models with some original models, including Xception, ResNet18 and VGG16-BN. The comparison results are shown in Table~\ref{table1}.
The IFMD models are obtained by embedding the FW module into the original models at different layers, such as the low layers, the middle layers, and the high layers.
In Table~\ref{table1}, None means the original models, and FW-L denotes that the IFMD model with the low layers' FW module. Similarly, FW-M and FW-H are the IFMD models at the middle and high layers. Comparing with the original models, we find that the IFMD models have small differences with respect to the accuracy. The FW module hardly affects the detection accuracy on different deep models, which means the feature decorrelation and constraint will not damage the detection performance. Our IFMD approach has reached a great performance on the face manipulated detection task.

\begin{figure*}[t]
\vskip 0.2in
\begin{center}
\centerline{\includegraphics[width=2.08\columnwidth]{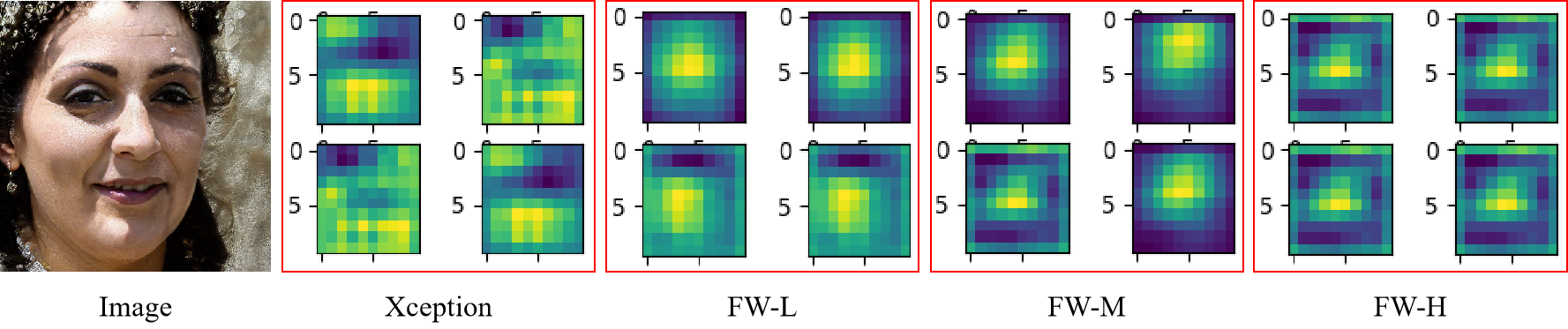}}
\caption{Comparisons of the IFMD models with the FW module embedded into different layers. We compare the feature maps of the second last layer from the IFMD models, obtained by embedding the FW module into the low (FW-L), middle (FW-M) and high (FW-H) layers. There are some differences in significance and interpretability for them, but they can all reach a better performance than Xception.}
\label{fig2}
\end{center}
\vskip -0.2in
\end{figure*}

\begin{figure}[t]
\vskip 0.02in
\begin{center}
\centerline{\includegraphics[width=\columnwidth]{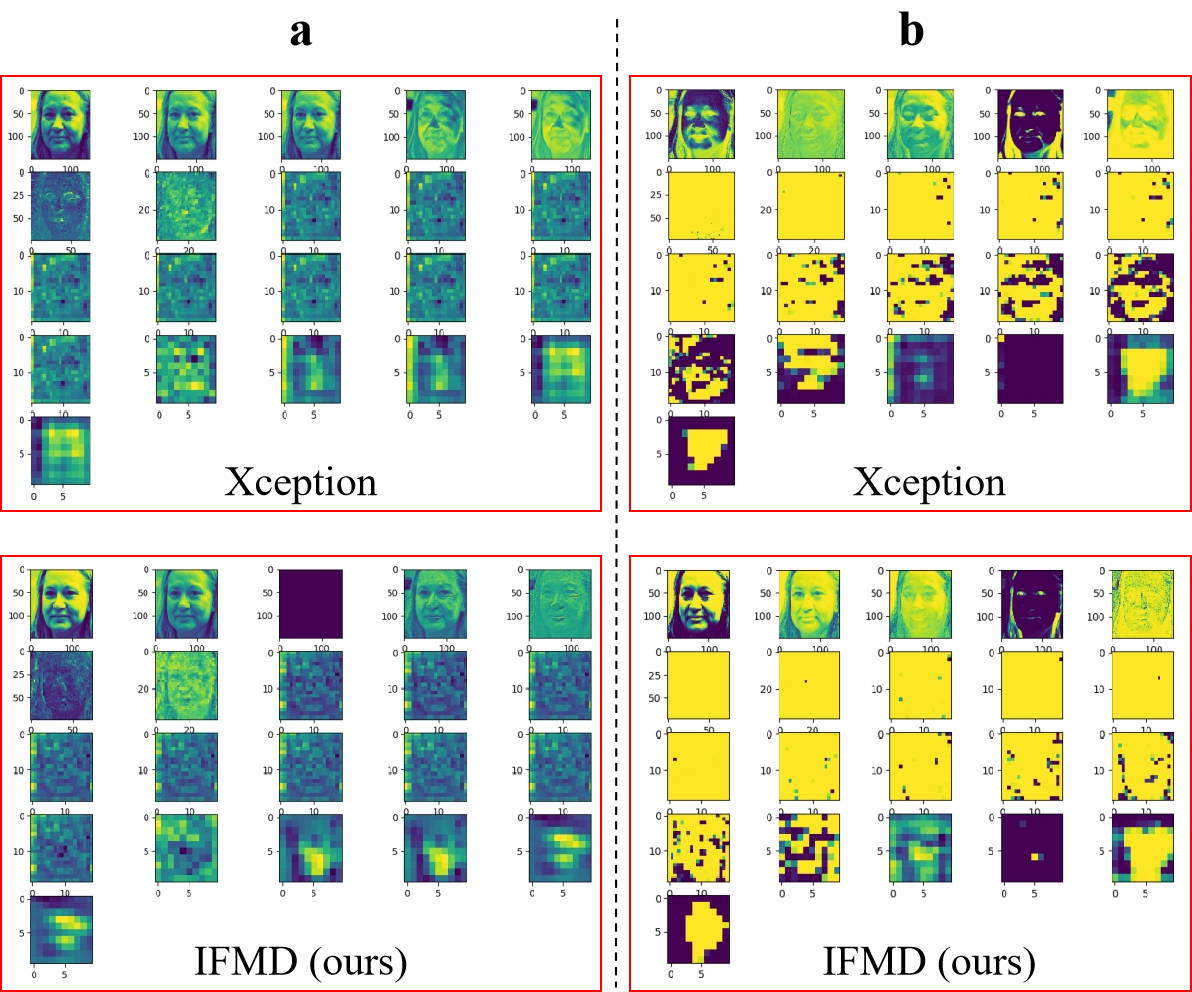}}
\caption{The feature maps of all the model layers for a manipulated face. \textbf{a}, the feature maps from one channel. \textbf{b}, the merged feature maps from all channels. The results of \textbf{a} show that the IFMD approach achieves better interpretability in the high-level semantic space of deep models. The results of \textbf{b} show that the IFMD approach does not lose the key features generally, resulting in equally good detection performance.}
\label{fig3}
\end{center}
\vskip -0.22in
\end{figure}

\balance
\subsection{Interpretability Performance}

To evaluate the interpretability performance, we visualize the internal feature maps to analyze the working mechanism based on the Xception model.
In Figure~\ref{fig1}, we show the feature maps of different channels in the second last layer for both fake and real images.
This figure shows that the feature representations of our IFMD model are more salient to represent the input, which demonstrates the FW module can distribute the internal features across different channels to reduce the representation complexity. It is easier for people to understand the high-level semantic information inside the IFMD model than the original Xception model.

In addition, we compare the internal feature representations of different IFMD models with the FW module embedded into different layers, such as FW-L, FW-M and FW-H. As shown in Figure~\ref{fig2}, we find that there are some differences in the interpretability performance due to the locations of the embedded FW module inside deep networks. However, all of them can achieve a better performance than the original models.
The IFMD approach improves the model interpretability by reducing the complexity and increasing the readability for the high-level information.


\subsection{Comprehensive Analysis}

In order to demonstrate that our IFMD model can strike a balance between the accuracy and interpretability, we further analyze the internal representations during the detection process.
As shown in Figure~\ref{fig3}, we visualize the features of one channel in the left subfigure (\textbf{a}) and the merged features of all the channels in the right subfigure (\textbf{b}).
From this figure, we find that the high-level features of the IFMD model are easier to understand from the perspective of people.
When we merge the features from all the channels, there is almost no difference between the Xception model and our IFMD model.
This means the FW module maintains the detection performance without losing the key information contributed to the inference, which further
explain why the detection accuracy of the IFMD model is hardly affected by the embedded FW module.

From the above experimental results, we find that our IFMD approach using the FW module can provide explanations on how the deep model comes to a particular prediction and why should we trust the detection results by whitening the internal inference process.
The IFMD approach makes it come true to balance the accuracy and interpretability of face manipulation detection.

\section{Conclusion and Future Work}
In this paper, we propose an interpretable face manipulation detection approach via feature whitening.
Specially, we design a feature whitening module to achieve the feature decorrelation and constraint of deep models for better understanding.
The decorrelation operation can distribute the internal representations into different channels to reduce complexity.
And the constraint operation aims to map the decorrelated features towards the salient area of manipulated faces for the model readability.
In this way, the approach can strike a balance between the model interpretability and detection accuracy.
The experiments demonstrate the effectiveness of our approach, including inference and interpretability.
In future work, we will improve this whitening approach to achieve a universally inference model interpretability,
and conduct more complete experiments to verify our proposed approach on a wide variety of detection tasks.

\newpage

\section{Acknowledgements}
This work was partially supported by grants from the National Key Research and Development Plan (2020AAA0140001), National Natural Science Foundation of China (61772513), Beijing Natural Science Foundation (19L2040), and the project from Beijing Municipal Science and Technology Commission (Z191100007119002). Shiming Ge is also supported by the Youth Innovation Promotion Association, Chinese Academy of Sciences.

\nocite{langley00}

\bibliography{ref}
\bibliographystyle{icml2021}

\end{document}